# An Evolutionary Approach to Drug-Design Using a Novel Neighbourhood Based Genetic Algorithm

*Arnab Ghosh[1], Avishek Ghosh[1], Arkabandhu Chowdhury[1], Amit Konar[1]*

*[1]Department of Electronics and Tele-communication Engineering*

*Jadavpur University, Kolkata-700032, India*

*Email:arnabju90@gmail.com,avishek.ghosh38@gmail.com,*

**Abstract**: The present work provides a new approach to evolve ligand structures which represent possible drug to be docked to the active site of the target protein. The structure is represented as a tree where each non-empty node represents a functional group. It is assumed that the active site configuration of the target protein is known with position of the essential residues. In this paper the interaction energy of the ligands with the protein target is minimized. Moreover, the size of the tree is difficult to obtain and it will be different for different active sites. To overcome the difficulty, a variable tree size configuration is used for designing ligands. The optimization is done using a novel Neighbourhood Based Genetic Algorithm (NBGA) which uses dynamic neighbourhood topology. To get variable tree size, a variable-length version of the above algorithm is devised. To judge the merit of the algorithm, it is initially applied on the well known Travelling Salesman Problem (TSP).

## 1. Introduction:

A strategy in drug design is to find compounds that bind to protein targets that constitute active sites which sustain viral proliferation. The challenge is to predict accurately structures of the compounds (ligands) when the active site configuration of the protein is known [1]. The literature addresses the challenge using a novel Genetic Algorithm that uses ring parent topology to generate offspring. It is found that the algorithm gives better candidate solution than traditional Genetic Algorithm many existing variation of it.

Evolutionary computation is used to place functional groups in appropriate leaves of the tree structured ligand. The objective is to minimize the interaction energy between the target protein and the evolved ligand, thus leading to the most stable solution. In [1] a fixed tree structure of the ligand is assumed. However it is difficult to get a prior knowledge of the structure and for a given geometry, no unique solution is the best solution. So variable length structure is used in the paper. Depending upon the geometry of the active site, a ligand can have a maximum or a minimum length (denoted by $l_{max}$ and $l_{min}$). The length of the ligand lies in between these two values.

## 2. Genetic Algorithm:

Genetic Algorithms (GAs) [2-6] are search algorithms based on the mechanics of the natural selection process (biological evolution). The most basic concept is that optimization is based on evolution, and the "Survival of the fittest" concept. GAs have the ability to create an initial population of feasible solutions, and then recombine them in a way to guide their search to only the most promising areas of the state space. Each feasible solution is encoded as a chromosome (string) also called a genotype, and each chromosome is given a measure of fitness via a fitness (evaluation or objective) function. The fitness of a chromosome determines its ability to survive and produce offspring. A finite population of chromosomes is maintained. GAs use probabilistic rules to evolve a population from one generation to the next. The generations of the new solutions are developed by genetic recombination operators:

- *Biased Reproduction*: selecting the fittest to reproduce
- *Crossover*: combining parent chromosomes to produce children chromosomes
- *Mutation*: altering some genes in a chromosome.

- Crossover combines the "fittest" chromosomes and passes superior genes to the next generation.
- Mutation ensures the entire state-space will be searched, (given enough time) and can lead the population out of a local minima.

### 2.1. Neighbourhood Based Approach (NBGA) :

Firstly we create a random sequence pool. Parents are selected randomly from the sequence pool and a ring parent topology is developed (shown in figure 1, 2). Consecutive two parents in the ring go under crossover process and two offspring are generated. After that trio selection procedure is applied (figure 3).

Pseudo code of selection procedure:

$Population^t = \{g_i^t\}; \ i = [1, \max\_pop]$

$Mutant = \{m_i\} = mutation(\{g_i^t\});$

$g_i^t = select(g_i^t, m_i);$

$Parent = \{p_i\} = rand\_select(\{g_i^t\})$

$Son = \{s_i\};$

$(s_{i1}, s_{i2}) = crossover(p_i, p_{i+1});$

$g_i^{t+1} = select(p_i, s_{i1}, s_{i2});$
$Population^{t+1} = \{g_i^{t+1}\};$

Select is a function that selects a sequence on the basis of cost function. Sequence with minimum cost function is selected.

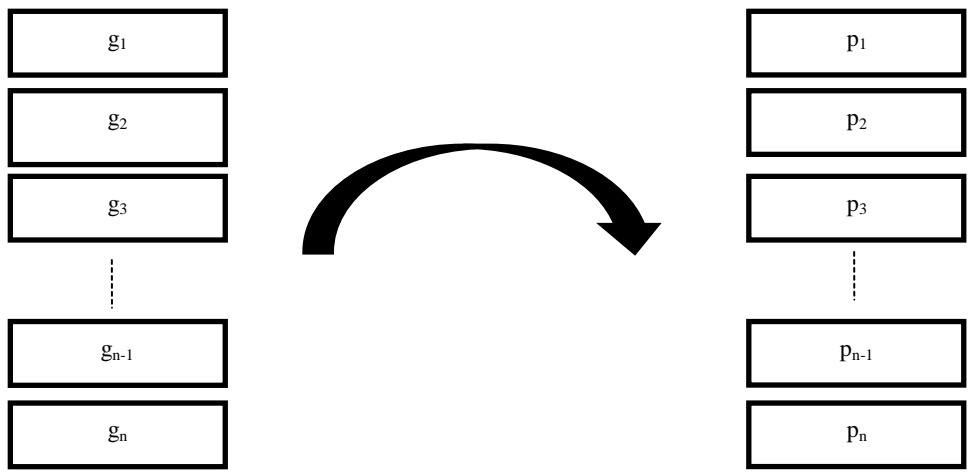

*Figure 1. Generation of parents after shuffling the population*

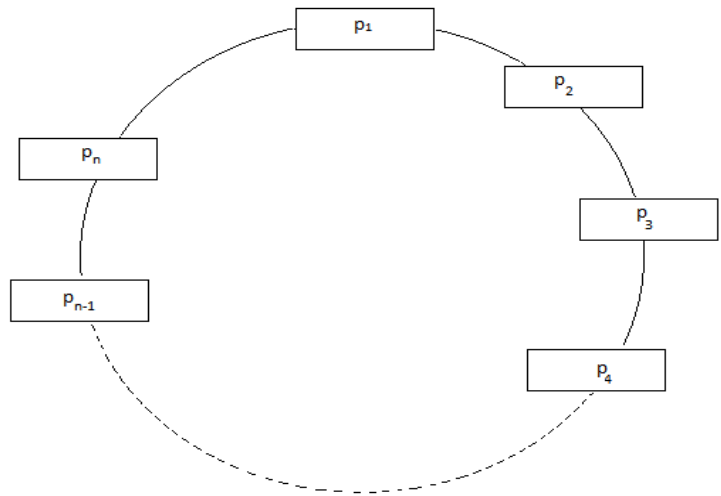

*Figure 2. Ring Parent Topology*

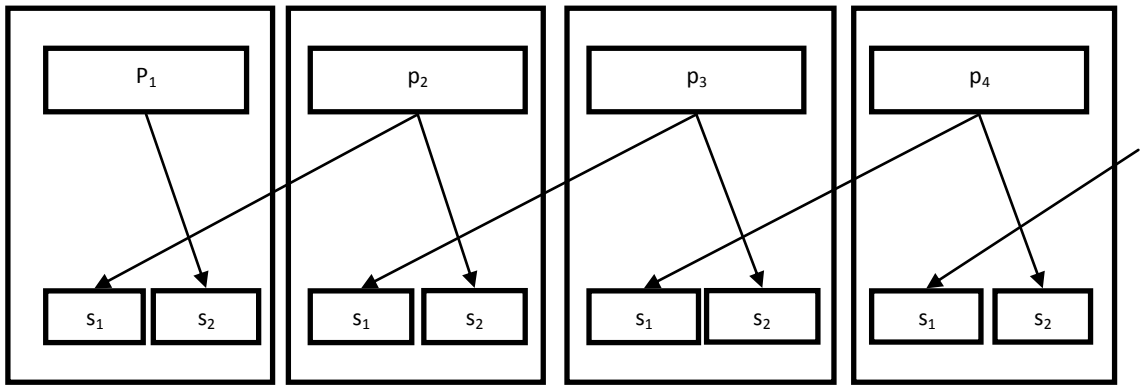

*Figure 3. Trio Selection*

## 2.2. Some Modified Mutation Schemes:

**Multiple Exchange mutation:** Here we select more than two dimensions at a time and exchange their position (i.e. randomly placed them at their place). This mutation ensures higher degree of convergence but accuracy becomes less. So we apply this scheme at the beginning of the algorithm and gradually drop out. Generally a random integer (ri) generator decides how many positions would be exchanged. Here we use highest number of position (hi) to be roughly one sixth of total no of dimensions and gradually decrease this number with generation ($2 < ri \leq hi$, $hi < n/6$, n=no of cities, gen=1). After a certain number of generation we use simple Exchange mutation technique (hi=2).

Illustrative example:

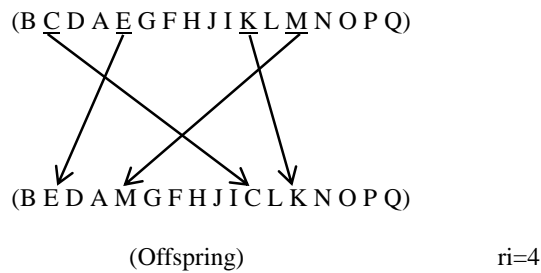

(Offspring)       ri=4

*Figure 4. Multiple Exchange Mutation*

**Multilevel mutation:** Actually two or more number of mutations are mixed up to generate offspring. There is no exact procedure to work out but we can apply same or another type of mutation in the offspring and thus increase the degree of mutation. This scheme will be effective based on the problem or situation. So very small probability is assigned and uses it after certain no of generation. We here use two multilevel mutations: exchange+displacement mutation and simple inversion+displacement mutation.

Second type is known as inversion mutation and can be very effective at the time of convergence.

Illustrative example:

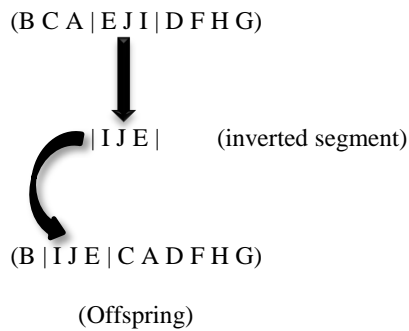

*Figure 5. Multilevel Mutation*

### 2.3. Pseudo Code:

**Begin** Neighbourhood Based GA

Create initial population

    **while** generation_count<k do

    /* k = max. number of generations. */

        **Begin**

        Mutation

        Select parent randomly

        Construct ring parent topology

        Crossover

        Apply trio selection procedure

        Increment generation_count

        **end**

    Output the best individual found

**end** Neighbourhood Based GA

### 3. Travelling Salesman Problem:

The *Travelling Salesman Problem* (TSP) is a well-known problem in the area of network and combinatorial optimization. Its importance stems from the fact that there is a plethora of fields in which it finds potential applications such as DNA fragment assembly and VLSI design. Formally, the TSP may be defined as a permutation problem with the objective of finding the path of the shortest length (or the minimum cost) on an undirected graph that represents cities or nodes to be visited. The traveling salesman starts at one node, visits all other nodes successively only one time each, and finally returns to the starting node. i.e., given n cities, named {c1, c2,....,cn}, and permutations, π1, ..., πn!, the objective is to choose πi such that the sum of all Euclidean distances between each node and its successor is minimized. The successor of the last node in the permutation is the first one. The Euclidean distance d, between any two cities with coordinate (x1, y1) and (x2, y2) is calculated by

$$d = \sqrt{(x1-x2)^2 + (y1-y2)^2} \tag{1}$$

More mathematically we may define the TSP as follows:

Given an integer $n \geq 3$ and an $n \times n$ matrix $C = (c_{ij})$, where each $c_{ij}$ is a nonnegative integer, which cyclic permutation π of the integers from 1 to *n* minimizes the sum $\sum_{i=1}^{n} c_{i\pi(i)}$ ?

The results are compared with those obtained using SWAP_GATSP [7], MMGA [8], IGA [9], OX SIM, (standard GA with order crossover and simple inversion mutation) [10] MOC SIM (Modified order crossover and SIM), and self organizing map (SOM) [11] for solving TSP. Table 1 summarizes the results obtained over 30 runs by running the NBGA, SWAP_GATSP, OX SIM and MOC SIM on the aforesaid nine different TSP instances

| Algo / Problems | Statistics | NBGA | SWAP_GATSP | OX_SIM | MOC_SIM |
|---|---|---|---|---|---|
| Grtschels24 24 1272 | Best Average Error | 1272 1272 0.0000 | 1272 1272 0.0000 | 1272 1272 0.0000 | 1272 1272 0.0000 |
| Bayg29 29 1610 | Best Average Error | 1610 **1610** **0.0000** | 1610 1615 0.3106 | 1610 1690 4.9689 | 1610 1622 0.7453 |
| Grtschels48 48 5046 | Best Average Error | 5046 **5084** **0.7531** | 5046 5110 1.2683 | 5097 5410 7.2136 | 5057 5184 2.7348 |
| eil51 51 426 | Best Average Error | **429** **432** **1.4084** | 439 442 3.7559 | 493 540 26.7606 | 444 453 6.3380 |
| St70 70 675 | Best Average Error | **682** **684** **1.3333** | 685 701 3.8519 | 823 920 36.2963 | 698 748 10.8148 |

*Table 1*: Comparison of NBGA with other Algorithms

Experimental results (Table 1) using NBGA are found to be superior in terms of quality of solution (best result, average result and error percentage) with less number of generations when compared with those of other existing GAs.

**4. Evolving Ligand Molecules:**

Protein with known active site configuration is used for evolving ligand structures. Our specific target is the known antiviral binding site of the Human Rhinovirus strain 14. This active site is known as the VP1 barrel for its resemblance with a barrel. The molecule which can easily be fit in the structure having minimum interaction energy will be the evolved drug (ligand). For simplification, a 2-dimensional structure is chosen [1].

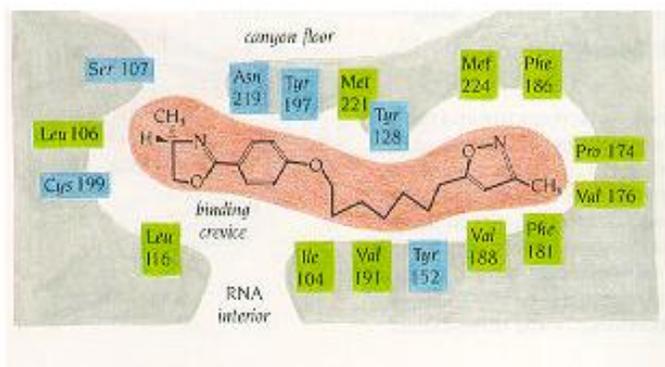

*Figure 6.* Active site of Human Rhinovirus strain 14

Figure 6 illustrates the binding site of Human Rhinovirus strain 14 and a typical structure of ligand is illustrated. For designing the ligand, the co-ordinates of the residues of the protein must be known. A ligand molecule is assumed to have a tree like structure on both sides of a fixed pharmacophore illustrated in figure 8. The structure has a left hand side known as left tree and a right hand side called right tree. For this configuration the right hand side contains 10 leaves and the left hand side contains 7 leaves. Each leaf denoted as a rounded number represents a functional group among the 8 functional groups listed in figure 7 along with their bond length projection on the x axis (table 2). It is to be noted that when group 8 occupies a position, the length of the tree is reduced enabling a variable length tree structure. For fixed length tree, group 8 is absent. The job is to find appropriate functional groups for the leaves such that the protein ligand docking energy (interaction energy) is minimized. Variable length NBGA is used to design variable size tree where fixed length structures are optimized with NBGA.

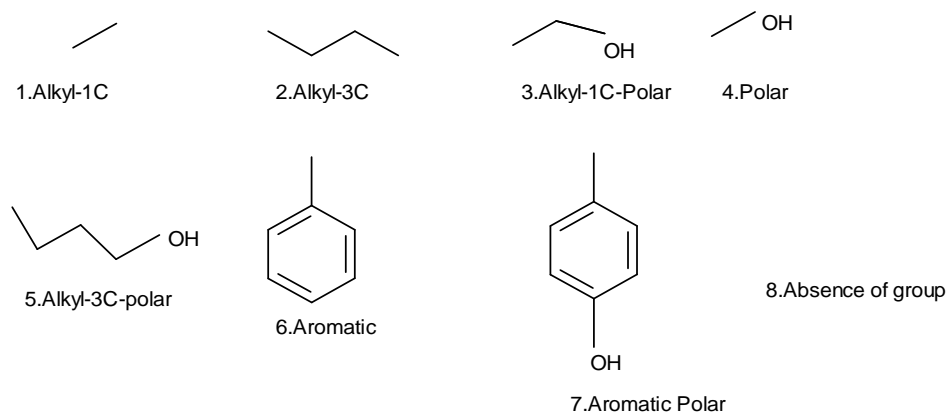

*Figure 7.* Functional Groups

| Functional group | Bond length along x axis (Å) | Code |
|---|---|---|
| Alkyl-1C | 0.65 | 1 |
| Alkyl-3C | 1.75 | 2 |
| Alkyl-1C-Polar | 1.1 | 3 |
| Alkyl-3C-Polar | 2.2 | 4 |
| Polar | 0.01 | 5 |
| Aromatic | 1.9 | 6 |
| Aromatic-Polar | 2.7 | 7 |
| NUL (no group) | - | 8 |

*Table 2.* Bond lengths of Functional Groups

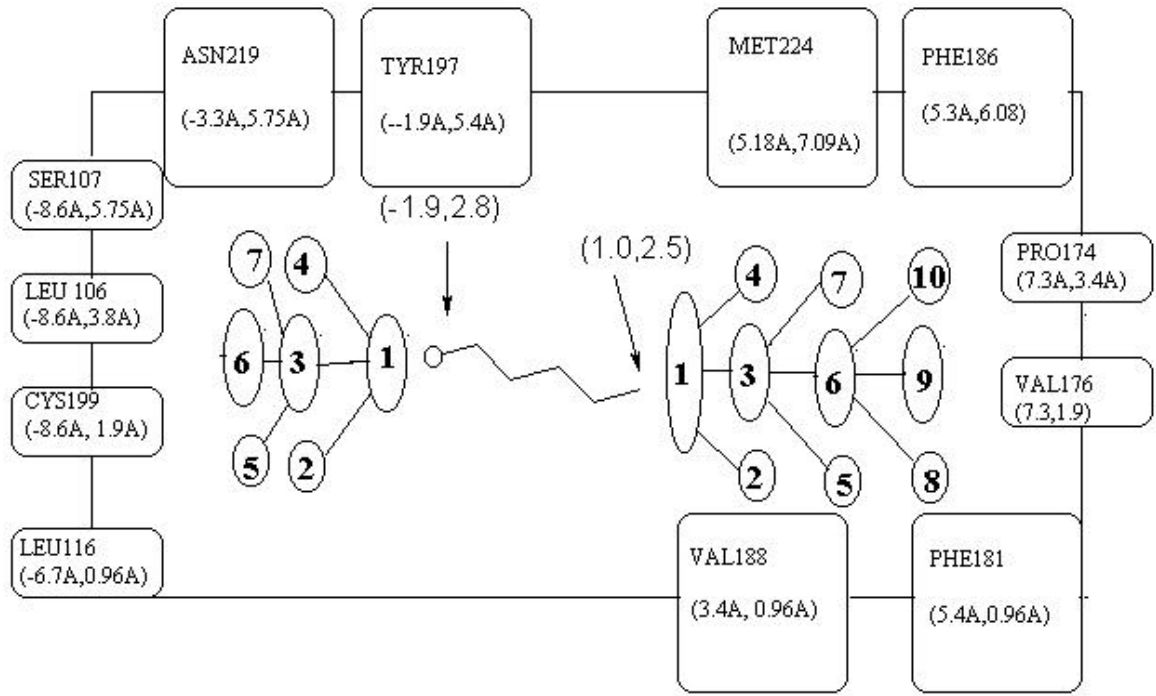

*Figure 8.* Active site co-ordinates of Human Rhinovirus strain 14

After computing the co-ordinates of each functional group, the euclidean distances between the group and all the residues of the active site are computed to find the Van der Waals energy.

**5. Variable-Length Version of NBGA:**

The size of representing tree is made variable. The length of the chromosome can vary between certain range denoted by $(l_{max}, l_{min})$, where these two are defined as:

$$l_{min} = \frac{length\ of\ the\ major\ axis}{maximum\ bond\ length}$$

$$l_{min} = \frac{length\ of\ the\ major\ axis}{minimum\ bondlength}$$

$l_{min}$ is calculated 7 for right tree and 2 for left tree. 8 is used to represent the absence of a functional group. One extra function named "Correct" is used to check correctness of the newly formed chromosome. Some constrains should be maintained. The 1st, 3rd and 6th position of right tree and 1st and 3rd position of left tree cannot contain any type of polar group unless they are the terminals of the tree. If any of the positions among 8th, 9th, 10th of the right tree contain any group then 6th position cannot contain 8. Similarly for left tree if any of the positions among 5th, 6th, and 7th contain any group then 3rd position cannot contain 8. Crossover, Mutation operators are implemented similarly as described above. We can look for a simple example: say right tree chromosome for two parents be:

[1 5 2 8| 8 1 5 |1 4 6]; [2 7 18 3 2 |5 8 6| 2]

Then corresponding two sons will be:

[1 5 2 8 5 8 6 1 4 6]; [2 7 18 3 2 8 1 5 2]

Now we can see that the right tree of the first son contain 8 at 6th position though its 8th, 9th and 10th places are not empty. So by using "Correct" function we place a 2 at 6th position so that the two sons will be:

[1 5 2 8 5 2 6 1 4 6]; [2 7 1 8 3 2 8 1 5 2]

These two chromosomes are valid and accepted.

### 6. Fitness Evaluation:

The computation of fitness was based on the interaction energy of the residues with the closest functional group and the chemical properties of these pairs. The distance between residues and functional groups should not be more than 2.7Å and less than 0.7Å. If the functional group and the closest residue are of different polarity, a penalty is imposed. The interaction energy of the ligand with the protein is the sum of Van der Waals potential energy between the groups and the amino acid residues present in the active site of the target protein. The Van der Waals potential energy is computed as :

$$V(r) = [(\frac{C_n}{r^6}) - (\frac{C_m}{r^{12}})]$$

Where $n$ and $m$ are integers and $C_n$ and $C_m$ are constants [12]. Finally fitness $F$ is computed as

$$F = \frac{k}{E}$$

Where $k$ is a constant (typically 100) and E is the total interaction energy in Kcal/mol. Therefore, maximizing fitness leads to minimizing interaction energy.

### 7. Results:

The initial population size is taken to be 100 and the algorithm is run for 100 generations. The interaction energy using Binary PSO (BPSO), Classical GA, Neighbourhood Based GA (NBGA), Variable length BPSO (VBPSO), Variable length NBGA (VNBGA) is tabulated in table 3. The configuration of the best right tree and the best left tree for both fixed length (using NBGA) and variable length (using VNBGA) is drawn in the figure 9 and figure 10. It is found that variable length structures lead to more stable and fit candidate solution than the fixed length structure. The fitness of best candidate solution is plotted against generation ($k$ is taken 100) in figure 11.

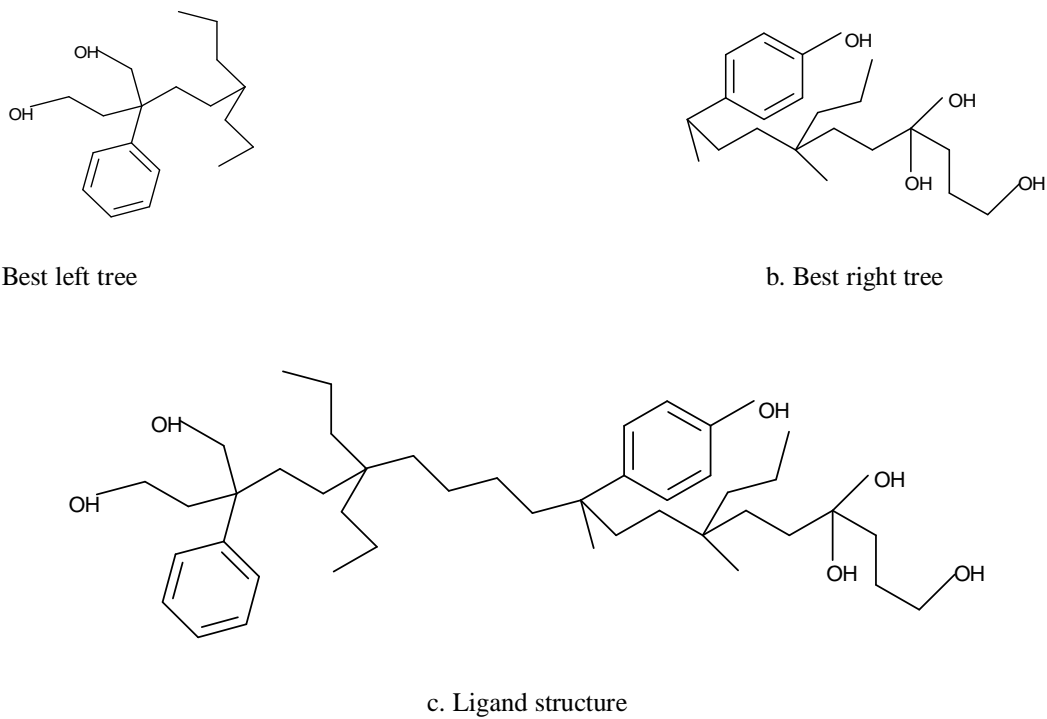

a. Best left tree

b. Best right tree

c. Ligand structure

*Figure 9*. Fixed length tree

**Variable Length Tree:**

Best left tree:

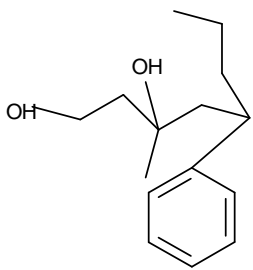

a. Best left tree

Best right tree:

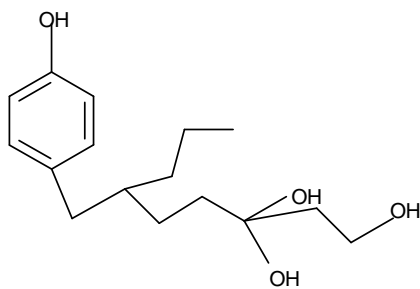

b. Best right tree

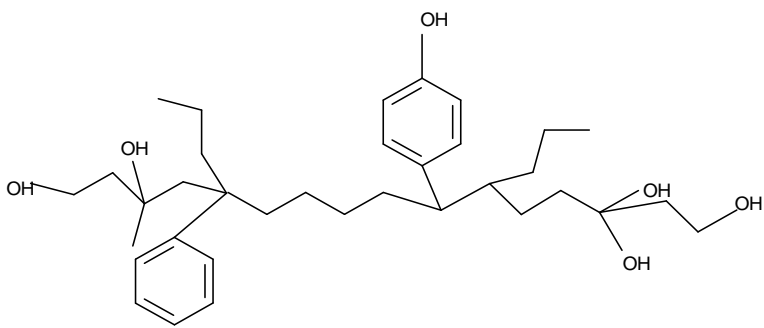

c. Ligand structure

*Figure 10.* Variable length structure

| Algorithm | Interaction energy (Kcal/mol) |
|---|---|
| Binary PSO (fixed length tree) | 13.76375 |
| GA (fixed length tree) | 11.64542 |
| NBGA (fixed length tree) | 11.57486 |
| VBPSO (variable length tree) | 8.57756 |
| VNBGA (variable length tree) | 8.10467 |

*Table 3.* Interaction energy values corresponding to Human Rhinovirus strain 14

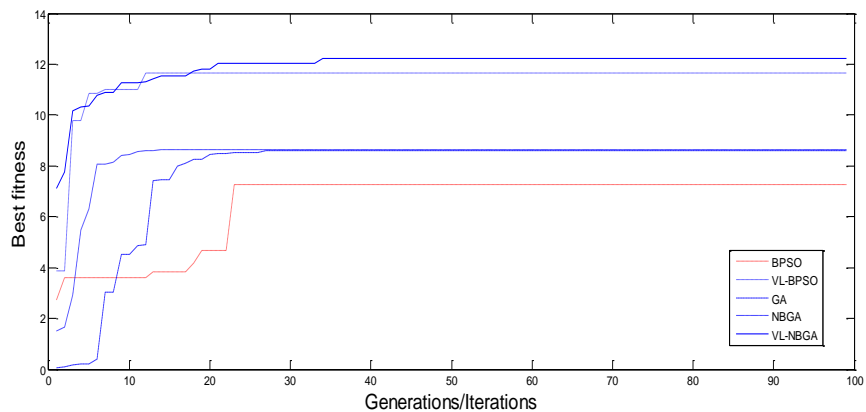

*Figure 11.* Computation of fitness over generations

## 8. Conclusions and Further research scopes:

We can conclude that Neighbourhood Based GA approach gives better result both for TSP and ligand design problem. It can be found that using VNBGA we can obtain more stable structure for ligand molecule. The paper uses a 2-dimensional approach which is quite unrealistic. A 3-dimensional approach using more complex tree structure will be our future research goal. Although protein-ligand interaction energy is minimized in the current paper, it may be required in some cases to minimize ligand energy to find stable solution. A multi-objective approach to minimize protein ligand interaction energy as well as ligand energy will be our future endeavor.